\documentclass{article}
\usepackage{ijcai25}

\usepackage{times}
\usepackage{soul}
\usepackage{url}
\usepackage[hidelinks]{hyperref}
\usepackage[utf8]{inputenc}
\usepackage[small]{caption}
\usepackage{subcaption}
\usepackage{graphicx}
\usepackage{amsmath}
\usepackage{amsthm}
\usepackage{booktabs}
\usepackage{algorithm}
\usepackage{algorithmic}
\usepackage[switch]{lineno}

\newcommand\blfootnote[1]{%
  \begingroup
  \renewcommand\thefootnote{}\footnotetext{#1}%
  \endgroup
}

\title{Learning What Matters: Causal Time Series Modeling\\for Arctic Sea Ice Prediction}

\author{
Emam Hossain
\and
Md Osman Gani
\affiliations
Department of Information Systems, University of Maryland Baltimore County, USA\\
\emails
\{emamh1, mogani\}@umbc.edu
}

\begin{document}

\maketitle

\blfootnote{\small Accepted and presented at the AI4TS Workshop @IJCAI 2025}

\begin{abstract}
Conventional machine learning and deep learning models typically rely on correlation-based learning, which often fails to distinguish genuine causal relationships from spurious associations, limiting their robustness, interpretability, and ability to generalize. To overcome these limitations, we introduce a causality-aware deep learning framework that integrates Multivariate Granger Causality (MVGC) and PCMCI+ for causal feature selection within a hybrid neural architecture. Leveraging 43 years (1979–2021) of Arctic Sea Ice Extent (SIE) data and associated ocean-atmospheric variables at daily and monthly resolutions, the proposed method identifies causally influential predictors, prioritizes direct causes of SIE dynamics, reduces unnecessary features, and enhances computational efficiency. Experimental results show that incorporating causal inputs leads to improved prediction accuracy and interpretability across varying lead times. While demonstrated on Arctic SIE forecasting, the framework is broadly applicable to other dynamic, high-dimensional domains, offering a scalable approach that advances both the theoretical foundations and practical performance of causality-informed predictive modeling.
\end{abstract}

\section{Introduction}

Machine learning (ML) and deep learning (DL) have significantly advanced predictive modeling across a wide range of domains, demonstrating strong capabilities in extracting intricate patterns from large-scale data. Despite these successes, most ML/DL models are grounded in correlation-based learning, which introduces fundamental limitations. Specifically, these models often fail to distinguish spurious correlations from genuine causal relationships, thereby reducing their robustness, interpretability, and generalization to unseen settings~\cite{pearl2018book}. For instance, although such models can uncover statistical regularities in the training data, they tend to break down when applied to new environments where the underlying causal structure has shifted.

Causal reasoning offers a compelling solution to this problem by uncovering the underlying mechanisms that drive system behavior. Unlike purely statistical approaches, causal discovery algorithms—such as Multivariate Granger Causality (MVGC)~\cite{barnett2014mvgc} and PCMCI+~\cite{runge2020discovering}—aim to identify both direct and indirect drivers of change, allowing models to focus on features with true explanatory power. MVGC, which generalizes the original Granger Causality method~\cite{granger1969investigating}, is well-suited to multivariate systems and offers scalability for high-dimensional time series. When incorporated into ML/DL models, causal features can enhance predictive reliability, improve model interpretability, and support generalization across diverse scenarios.

\begin{figure}[!h]
\centering
\includegraphics[width=0.9\columnwidth]{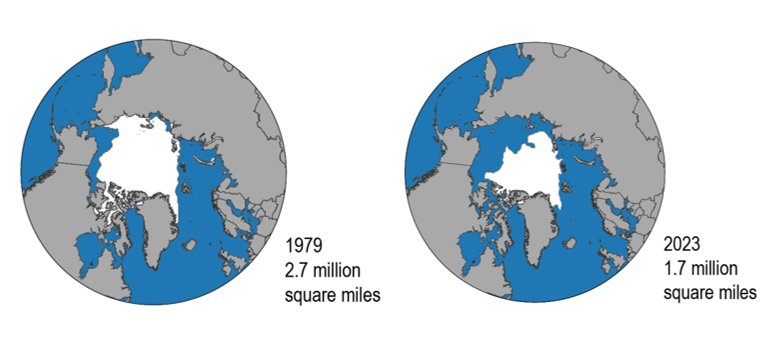}
\caption{Approximately 38\% decline in Arctic September sea ice, from 2.7 million square miles in 1979 to 1.7 million square miles in 2023 (Source: US Global Change Research Program).}
\label{fig:arctic_trends}
\end{figure}

These issues are particularly evident in dynamic, high-dimensional domains such as climate modeling. A prime example is the prediction of Arctic sea ice extent (SIE), which depends on complex, nonlinear interactions among atmospheric and oceanic variables. The accelerating loss of Arctic sea ice (Figure~\ref{fig:arctic_trends}) carries serious consequences for global weather systems, ecosystems, and human infrastructure. Yet traditional forecasting techniques—along with correlation-based ML/DL approaches—continue to struggle with long-term SIE prediction, largely due to their inability to capture the intricate causal dependencies inherent in the Arctic climate system~\cite{andersson2021seasonal}.

To bridge these limitations, this study proposes a causality-guided deep learning framework for forecasting Arctic Sea Ice Extent (SIE), integrating causal discovery with temporal neural architectures. Leveraging 43 years (1979–2021) of ocean-atmospheric data, we apply Multivariate Granger Causality (MVGC) and PCMCI+ to identify a refined set of causally relevant predictors. These features are then used as inputs to a hybrid neural model composed of Gated Recurrent Units (GRUs) and Long Short-Term Memory (LSTM). This design enables the model to emphasize variables with direct influence on sea ice behavior, reduce unnecessary input complexity, and improve computational efficiency. By constraining the model to learn from causal drivers rather than all correlated signals, the approach enhances prediction robustness and interpretability while simplifying the learning process.

The primary contributions of this work are as follows: (1) We apply MVGC and PCMCI+ to uncover causal dependencies between Arctic SIE and ocean-atmospheric variables, offering interpretable insights into the mechanisms of change. (2) We design a hybrid GRU-LSTM model that incorporates these causal features to improve both short- and long-range forecasting performance. (3) We conduct a comprehensive empirical evaluation using RMSE, MAE, and \(R^2\), demonstrating that causality-informed modeling significantly improves predictive reliability for lead times extending up to six months. In doing so, this work contributes a practical and interpretable bridge between causality and deep learning-based climate forecasting.

\section{Related Works}

This section reviews major developments in Arctic sea ice forecasting, focusing on machine learning, deep learning, and causality-based modeling approaches.

\subsection{Machine Learning in Arctic Sea Ice Prediction}

Machine learning (ML) approaches have been extensively employed for Arctic sea ice prediction due to their capacity to handle large-scale data and model complex nonlinear relationships. \cite{zhu2024stdnet} introduced the Spatio-Temporal Decomposition Network (STDNet), which improves the accuracy of sea ice concentration forecasts. \cite{driscoll2024data} developed a data-driven emulator for melt pond prediction by integrating physical insights with ML models. In a similar direction, \cite{koo2024hierarchical} proposed a hierarchical convolutional neural network (CNN) that combines multiple ice-related indicators to enhance forecasting accuracy.

Deep learning (DL) methods have further advanced the field by capturing intricate spatial-temporal dependencies. \cite{xu2024sifm} applied a foundational DL architecture for multi-resolution sea ice forecasting, while \cite{ren2024sicnet} leveraged transformer-based models to incorporate sea ice thickness in seasonal projections. \cite{kim2025long} demonstrated the effectiveness of U-Net-based frameworks by integrating climate drivers such as surface temperature and radiation into long-term forecasting pipelines. Additionally, \cite{liu2024physics} proposed a physics-informed DL model capable of predicting both sea ice concentration and velocity.

While these approaches have yielded notable progress, their reliance on statistical correlations remains a key limitation. Such dependency can lead to model overfitting and hinder interpretability, particularly in dynamic systems like the Arctic climate~\cite{dunmireSGL2025}.

\subsection{Causality-Driven Predictive Modeling}

In light of the shortcomings of correlation-based learning, recent studies have increasingly explored the integration of causal reasoning into ML/DL models. For example, \cite{oliveira2024causality} embedded causal knowledge into financial time series forecasting, resulting in improved generalization and interpretability. Similarly, \cite{li2024advancing} employed Granger Causality in the context of Arctic sea ice prediction, showcasing its potential in uncovering dynamic dependencies within environmental systems.

Building upon these developments, our study combines Multivariate Granger Causality (MVGC) and PCMCI+ with a hybrid GRU-LSTM network to enhance Arctic SIE forecasting. By focusing on causally relevant ocean-atmospheric drivers, the model minimizes redundant input features and achieves stronger performance for both short- and long-range prediction tasks. Our experimental results affirm that this causality-informed approach is well-suited to the complexity of high-dimensional systems like the Arctic climate, underscoring the value of causal inference in building robust and interpretable predictive models.

\section{Background}

This section introduces the foundational components of our approach: time series causal discovery algorithms and recurrent neural networks (RNNs), which are central to time series modeling.

\subsection{Time Series Causal Discovery}

\begin{figure}[h!]
\centering
\includegraphics[width=0.9\columnwidth]{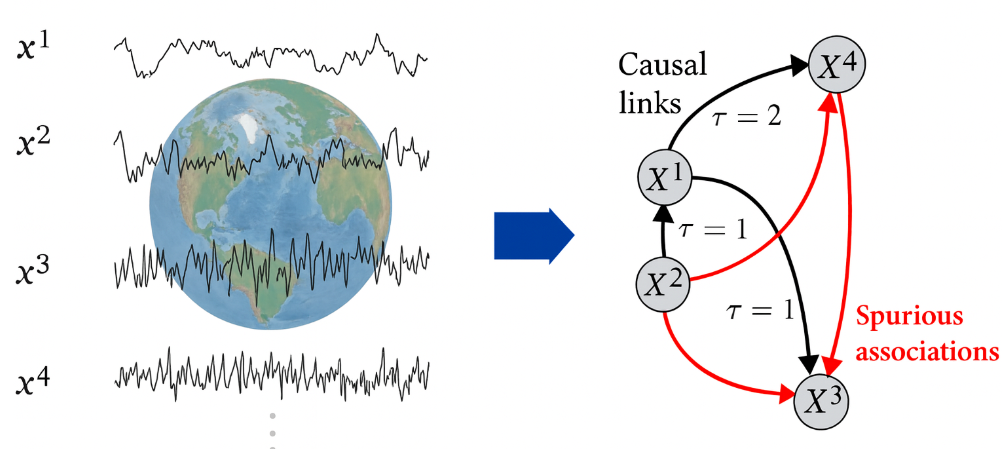}
\caption{Illustration of how causal discovery algorithms identify lagged causal relationships from time series data. $\tau$ represents the timelag of the causal links.}
\label{fig:causal_discovery_process}
\end{figure}

Causal discovery aims to reveal underlying cause-and-effect structures within time series data, offering insights that go beyond surface-level statistical correlations~\cite{hasansurvey}. From observational data, these methods construct a causal graph (Figure~\ref{fig:causal_discovery_process}) that captures both the directional dependencies among variables and their associated time lags~\cite{ferdous2025timegraph}. Such representations are particularly valuable in predictive modeling for complex systems, enhancing interpretability, generalizability, and robustness—especially when nonlinear interactions are involved.

\begin{figure*}[!h]
\centering
\includegraphics[width=1.8\columnwidth]{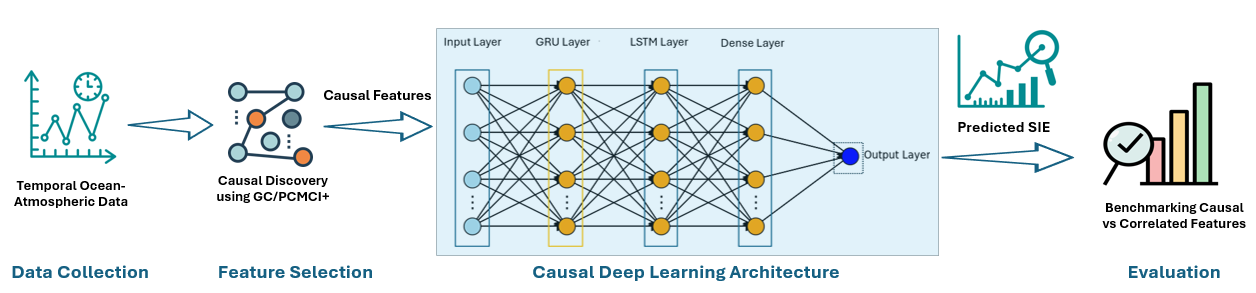}
\caption{Overview of the proposed causal deep learning framework for Arctic Sea Ice Extent (SIE) prediction. Daily and monthly temporal ocean-atmospheric datasets are analyzed with MVGC and PCMCI+ to identify causal features. These features are integrated into a GRU-LSTM architecture for SIE forecasting, and performance is evaluated on causal vs. correlated features.}
\label{fig:methodology}
\end{figure*}

\textbf{Multivariate Granger Causality (MVGC)}~\cite{barnett2014mvgc}, a generalization of the original Granger Causality (GC) method~\cite{granger1969investigating}, evaluates whether the inclusion of lagged values from multiple time series enhances the prediction of a target variable. Unlike standard GC, MVGC is tailored for multivariate contexts and scales well to high-dimensional data. Its efficiency and ability to model inter-variable interactions make it particularly suitable for environmental systems such as the Arctic, where numerous variables interact over time.

\textbf{PCMCI+}~\cite{runge2020discovering} builds upon the PC (Peter-Clark) algorithm by incorporating momentary conditional independence (MCI) tests. It is specifically designed to address challenges posed by autocorrelation and high dimensionality, allowing it to disentangle direct causal influences from indirect ones. PCMCI+ is especially effective in identifying meaningful causal relationships in environmental datasets that are often plagued by noise and spurious correlations~\cite{hossain2025correlation,hossain2024incorporating}.

\subsection{Recurrent Neural Networks}

Recurrent Neural Networks (RNNs), including variants such as Gated Recurrent Units (GRU) and Long Short-Term Memory (LSTM) networks, are extensively used for time series forecasting due to their capacity to model sequential dependencies \cite{dey2021comparative}. These architectures are designed to overcome issues like the vanishing gradient problem, enabling the learning of both short-range and long-range temporal patterns.

\textbf{Gated Recurrent Unit (GRU)}~\cite{cho2014learning} offers a streamlined alternative to traditional RNNs by integrating the update and reset gates into a simplified structure. This reduction in parameter complexity allows GRUs to deliver computational efficiency without sacrificing predictive performance. In contrast, \textbf{Long Short-Term Memory (LSTM)} networks~\cite{hochreiter1997long} utilize memory cells along with input, output, and forget gates to retain information over extended time intervals. LSTMs have proven particularly effective in forecasting tasks involving environmental and climate data~\cite{liu2024physics}.

By combining the strengths of both architectures, hybrid GRU-LSTM models leverage the speed and efficiency of GRUs with the memory retention capabilities of LSTMs. Prior studies have shown that such hybrid models outperform their standalone counterparts in various forecasting scenarios~\cite{islam2021foreign,hossain2020novel}, making them well-suited for incorporating causally selected features in predictive modeling pipelines.

\section{Methodology}

This section details the proposed causality-guided framework for forecasting Arctic Sea Ice Extent (SIE). The framework comprises four primary stages: (a) data acquisition and preprocessing, (b) causal variable selection, (c) development of a causal deep learning architecture, and (d) model training and evaluation. The overall pipeline is depicted in Figure~\ref{fig:methodology}.

\subsection{Data Collection and Preprocessing}

We utilize a combination of oceanic and atmospheric datasets, alongside sea ice extent (SIE) records, to explore both long-term trends and seasonal variability within the Arctic climate system~\cite{ali2021sea}. The data originate from two main sources: ERA-5 reanalysis products supply the ocean-atmospheric variables, while sea ice extent values are derived from passive microwave observations (Nimbus-7 SSMR and DMSP SSM/I-SSMIS), as curated by the National Snow and Ice Data Center (NSIDC)~\cite{cavalieri1996sea}.

\begin{table}[h!]
\centering
\caption{Summary of Daily \& Monthly Sea Ice Datasets.}
\label{tab:variables_summary}
\begin{tabular}{@{}lll@{}}
\toprule
\textbf{Variables}          & \textbf{Range}           & \textbf{Unit}       \\ \midrule
Surface Pressure           & [400, 1100]             & hPa                 \\
Wind Velocity              & [0, 40]                 & m/s                 \\
Specific Humidity          & [0, 0.1]                & kg/kg               \\
Air Temperature            & [200, 350]              & K                   \\
Shortwave Radiation        & [0, 1500]               & W/m\textsuperscript{2} \\
Longwave Radiation         & [0, 700]                & W/m\textsuperscript{2} \\
Rainfall                   & [0, 800]                & mm/day              \\
Snowfall                   & [0, 200]                & mm/day              \\
Sea Surface Temperature    & [200, 350]              & K                   \\
Sea Surface Salinity       & [0, 50]                 & psu                 \\
Sea Ice Extent             & [3.34, 16.63]           & Million             \\ \bottomrule
\end{tabular}
\end{table}

\begin{table*}[!h]
\centering
\caption{Input features used for the trained models.}
\label{tab:model_inputs}
\resizebox{2\columnwidth}{!}{%
\begin{tabular}{@{}lll@{}}
\toprule
\textbf{Model} & \textbf{Datasets Used} & \textbf{Input Features} \\ \midrule
$DL_{\text{vanilla}}$  & Daily / Monthly & All 10 ocean-atmospheric variables \\
$DL_{\text{GC}}$  & Daily / Monthly & \textit{Surface Pressure, Wind Velocity, Specific Humidity, Air Temperature, Shortwave Radiation, Longwave Radiation, Rainfall, Snowfall, SSS, SIE} \\
$DL_{\text{PCMCI+}}$  & Daily & \textit{Surface Pressure, Longwave Radiation, Snowfall, SSS, SIE} \\
$DL_{\text{PCMCI+}}$  & Monthly & \textit{Longwave Radiation, SST, SIE} \\
$DL_{\text{DPCMCI+}}$  & Monthly & \textit{Surface Pressure, Longwave Radiation, Snowfall, SSS, SIE} \\ \bottomrule
\end{tabular}%
}
\end{table*}

To capture comprehensive climate behavior, we construct two temporally distinct time series datasets. The first consists of monthly gridded measurements, spatially aggregated over the Arctic region north of 25°N using area-weighted averaging, spanning the years 1978 to 2021. The second dataset contains daily gridded records for the same spatial domain, covering the period from 1979 to 2018, thus supporting the identification of short-term fluctuations and lagged causal interactions. In total, the dataset includes ten ocean-atmospheric predictors and sea ice extent values, as summarized in Table~\ref{tab:variables_summary}.

To ensure data consistency and analytical validity, we apply a series of preprocessing steps: normalization, temporal aggregation, and imputation of missing values. These preprocessing procedures minimize data noise and prepare the time series for downstream causal analysis, facilitating the discovery of key drivers underlying Arctic sea ice variability.

\subsection{Causal Feature Identification}


Identifying causally relevant features is a critical step toward improving both interpretability and predictive skill in Arctic SIE forecasting. In this work, we systematically apply the MVGC and PCMCI+ algorithms to uncover the most influential drivers of sea ice variability. For both the daily and monthly datasets, \textbf{MVGC} revealed that all variables, with the exception of \textit{Sea Surface Temperature (SST)}, exert a significant causal influence on Arctic sea ice (Table \ref{tab:model_inputs}). These results demonstrate the widespread causal impact of atmospheric and oceanic processes on SIE dynamics and provide a principled basis for feature selection in our modeling framework, ensuring greater robustness and clarity in subsequent analyses.

\begin{figure}[!h]
\centering
\begin{subfigure}{\linewidth}
    \centering
    \includegraphics[width=0.85\linewidth]{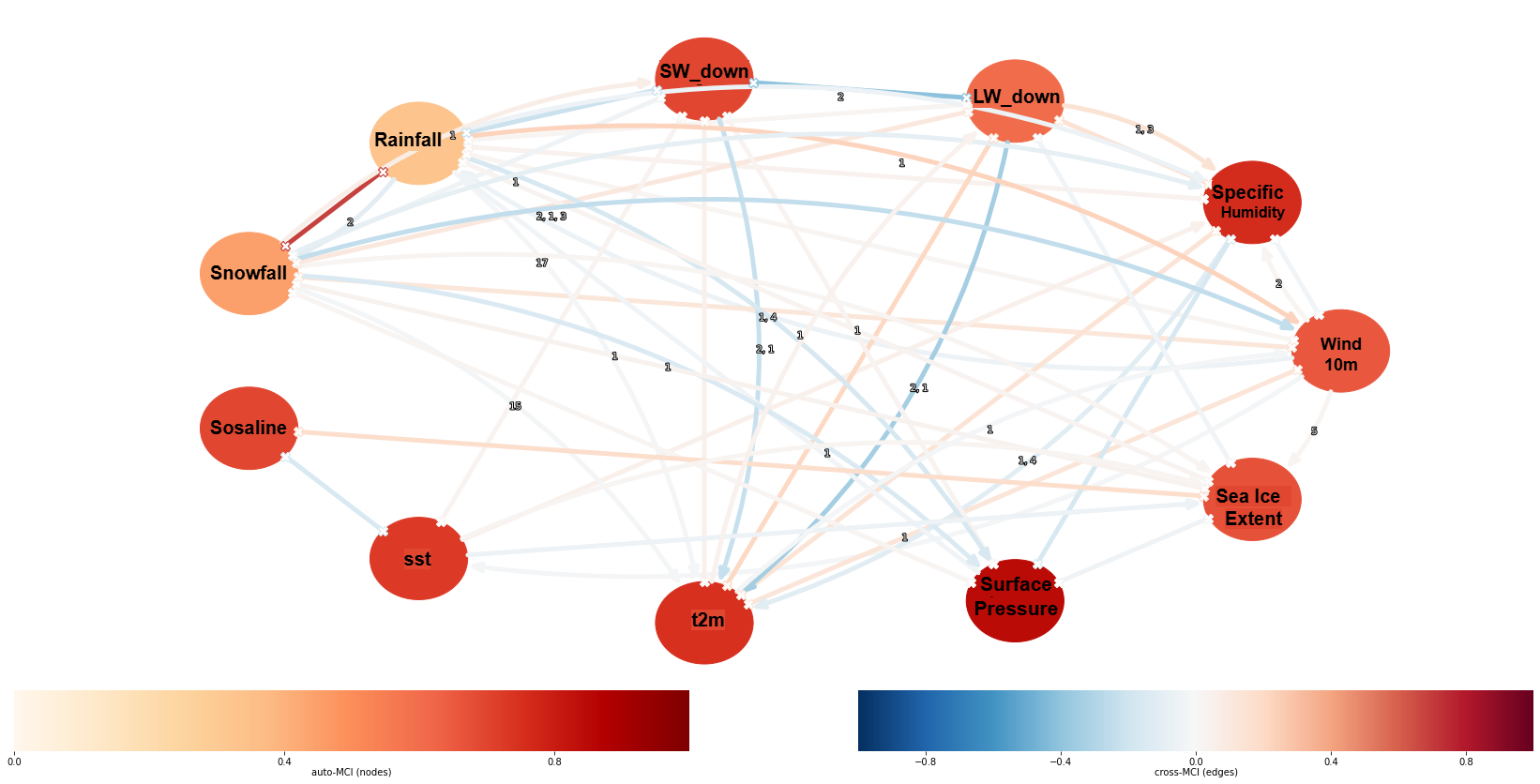}
    \caption{Causal graph of PCMCI+ for daily data.}
    \label{fig:causal_graph_daily}
\end{subfigure}
\vfill
\begin{subfigure}{\linewidth}
    \centering
    \includegraphics[width=0.85\linewidth]{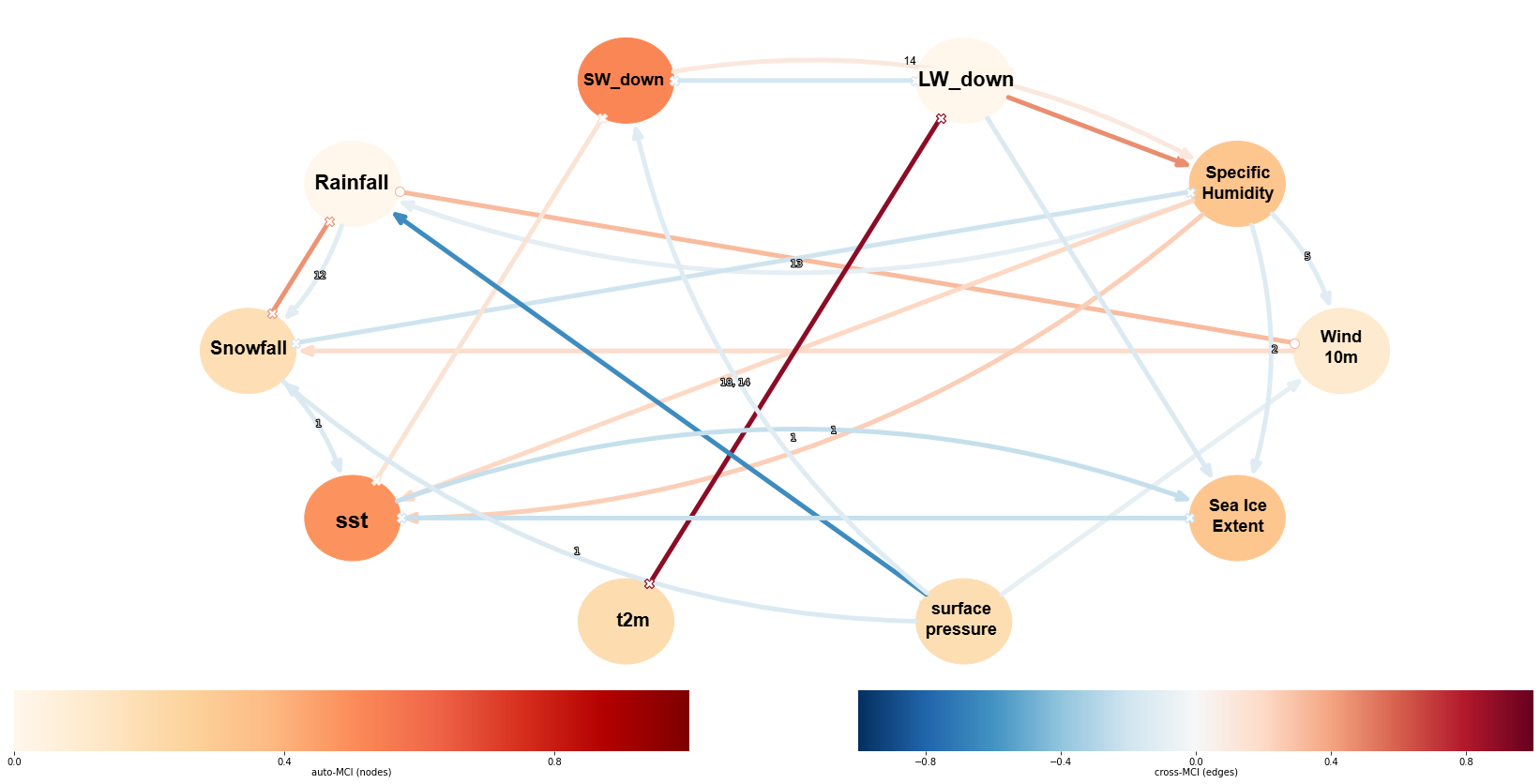}
    \caption{Causal graph of PCMCI+ for monthly data.}
    \label{fig:causal_graph_monthly}
\end{subfigure}
\caption{Causal graphs generated by PCMCI+ for (a) daily and (b) monthly datasets, illustrating the causal relationships between ocean-atmospheric variables and SIE.}
\label{fig:causal_graphs}
\end{figure}

\textbf{PCMCI+}, recognized for its capability to handle autocorrelation and high-dimensionality in time series data, yielded a more selective set of causal predictors. In the daily dataset, PCMCI+ identified \textit{longwave radiation, snowfall, sea surface salinity (SSS), surface pressure}, and \textit{SIE} itself as the dominant causal variables. In contrast, for the monthly dataset, the key causal features were limited to \textit{longwave radiation, SST}, and \textit{SIE} (Table \ref{tab:model_inputs}). These findings reflect potential temporal differences in causal mechanisms, suggesting that daily and monthly resolutions may capture distinct dynamics within the Arctic system.

Figure~\ref{fig:causal_graphs} presents the causal graphs produced by PCMCI+ for both daily and monthly datasets, depicting the direct causal influences of environmental variables on SIE. The features identified through this process were subsequently used as inputs to the GRU-LSTM model, ensuring that the learning algorithm focused on causally meaningful information for improved forecasting accuracy.

\begin{algorithm}[!h]
\caption{Causal Deep Learning Framework for Arctic Sea Ice Prediction}
\begin{algorithmic}[1]
\label{alg:arctic_sie}
\STATE \textbf{Input:} Multivariate time series $\mathcal{D} = \{X_t, Y_t\}$, where $X_t$ denotes ocean-atmospheric variables and $Y_t$ is the target SIE. Use maximum lag $\tau = 21$; apply MVGC and PCMCI+ for causal discovery.
\STATE \textbf{Output:} Predicted SIE values $\hat{Y}_{t+h}$ for forecast horizons $h \in \{1, \dots, 6\}$.

\STATE \textbf{Step 1: Preprocessing}
\STATE Retrieve and preprocess both daily and monthly datasets $\mathcal{D}_{\text{daily}}$ and $\mathcal{D}_{\text{monthly}}$ from ERA-5 and NSIDC archives.
\STATE Normalize inputs, impute missing values, and generate lagged versions of variables up to $\tau$.

\STATE \textbf{Step 2: Causal Feature Identification}
\STATE Use MVGC to extract causal variables $C_{\text{GC}} \subseteq X_t$.
\STATE Apply PCMCI+ to determine causal subsets $C_{\text{PCMCI+}} \subseteq X_t$ for both temporal resolutions.

\STATE \textbf{Step 3: Model Development and Training}
\STATE Define the GRU-LSTM model as follows:
\begin{itemize}
    \item \textbf{Input Layer:} features from $C \in \{X_t, C_{\text{GC}}, C_{\text{PCMCI+}}\}$.
    \item \textbf{GRU Layer:} $64$ units with dropout $p=0.2$.
    \item \textbf{LSTM Layer:} $128$ units with dropout $p=0.2$.
    \item \textbf{Dense Layer:} $64$ units for feature fusion.
    \item \textbf{Output Layer:} Single neuron for predicting $\hat{Y}_{t+h}$.
\end{itemize}

\STATE Train the following model variants:
\begin{itemize}
    \item $\mathbf{DL_{\text{vanilla}}}$ using the full input $X_t$.
    \item $\mathbf{DL_{\text{GC}}}$ using MVGC-derived features $C_{\text{GC}}$.
    \item $\mathbf{DL_{\text{PCMCI+}}}$ using PCMCI+-derived features $C_{\text{PCMCI+}}$.
    \item $\mathbf{DL_{\text{DPCMCI+}}}$ using daily-derived $C_{\text{PCMCI+}}^{\text{daily}}$ for monthly predictions.
\end{itemize}

\STATE Use the \textit{Adam} optimizer, mean squared error (MSE) loss, a batch size of $64$, and train for up to $100$ epochs with early stopping to prevent overfitting.

\STATE \textbf{Step 4: Model Evaluation}
\STATE Assess all models on the test set $\mathcal{D}_{\text{test}}$ using RMSE, MAE, and $R^2$ metrics across all lead times $h \in \{1, \dots, 6\}$.
\end{algorithmic}
\end{algorithm}

\subsection{Designing Causal Deep Learning Model}

To effectively utilize the set of causally identified predictors, we design a hybrid deep learning model that integrates Gated Recurrent Units (GRUs) and Long Short-Term Memory (LSTM) layers. This architecture is tailored to capture temporal dependencies at multiple scales—short-term patterns via GRUs and long-term trends via LSTMs. The input to the model consists of time series constructed with a lookback window of 21 timesteps, aligned with the maximum lag length used in the causal discovery process.

The model architecture includes an input layer with 21 neurons representing the historical sequence, followed by a GRU layer comprising 64 units and a 20\% dropout rate for modeling short-range dynamics. This is followed by an LSTM layer with 128 units and 20\% dropout to capture long-term dependencies. A fully connected dense layer with 64 neurons integrates learned representations, and the final output layer—consisting of a single neuron—predicts sea ice extent (SIE) at lead times ranging from 1 to 6 months (Figure~\ref{fig:methodology}).

\subsection{Model Training and Evaluation}

The GRU-LSTM models were trained using historical data up to the year 2013, reserving 10\% of the training split for validation. The test set comprised observations from 2014–2018 for the daily dataset and 2014–2021 for the monthly dataset. To assess the influence of causal feature integration, we trained separate models for both temporal granularities. Across both setups, we evaluated three core model variants: $DL_{\text{vanilla}}$, trained on the complete set of 10 ocean-atmospheric variables; $DL_{\text{GC}}$, trained on the subset identified via Multivariate Granger Causality; and $DL_{\text{PCMCI+}}$, which used features selected by PCMCI+. For the monthly data, we also included a fourth configuration, $DL_{\text{DPCMCI+}}$, which utilized PCMCI+ features discovered from daily data but repurposed for monthly forecasting. The input feature sets used for each model are summarized in Table~\ref{tab:model_inputs}. This setup enables a comprehensive comparison of how causal selection impacts the predictive performance of the deep learning models.

All models were trained using the \textit{Adam} optimizer, employing mean squared error (MSE) as the loss function. Training was performed with a batch size of 64 over a maximum of 100 epochs, with early stopping used to prevent overfitting. Model performance was assessed using three widely adopted metrics: Root Mean Squared Error (RMSE), which captures the average magnitude of prediction errors; Mean Absolute Error (MAE), which evaluates the average absolute deviation from actual values; and the Coefficient of Determination (\(R^2\)), which quantifies the proportion of variance explained by the model. The datasets and code used in this study are publicly accessible on GitHub\footnote{\url{https://github.com/ehfahad/learning_what_matters}}. A step-by-step overview of the complete workflow is outlined in Algorithm~\ref{alg:arctic_sie}.

\section{Results and Discussion}

\begin{figure}[h!]
\centering
\includegraphics[width=0.95\columnwidth]{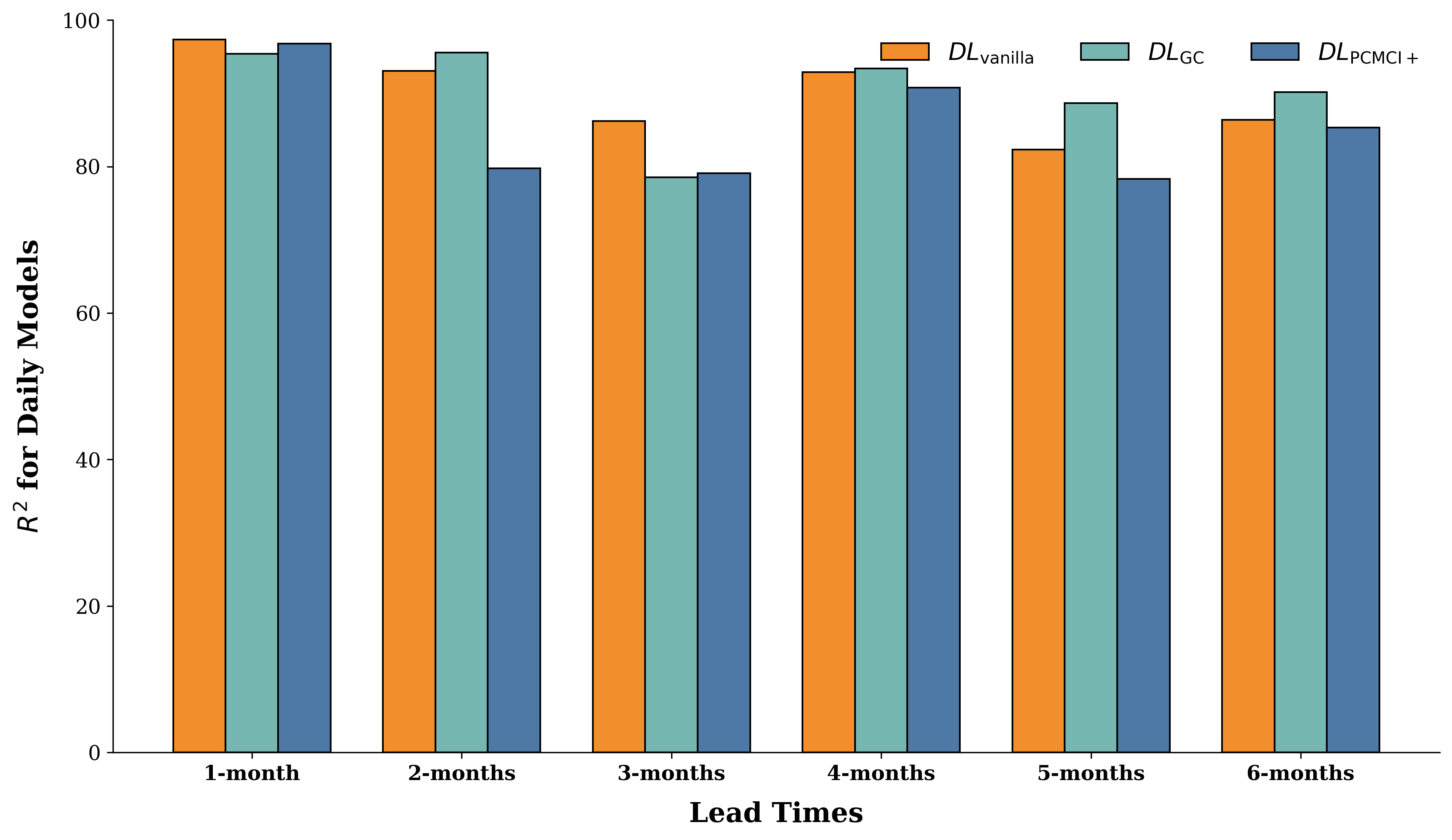}
\caption{\(R^2\) values for daily models across the lead times.}
\label{fig:R2_daily}
\end{figure}

This section presents an evaluation of the proposed causality-guided deep learning framework for forecasting Arctic Sea Ice Extent (SIE). Our analysis demonstrates how incorporating causally relevant features enhances both the predictive accuracy and interpretability of the model. Table~\ref{tab:daily_metrics_full} summarizes the RMSE and MAE values obtained from daily models trained using three different feature sets: the full set of ocean-atmospheric variables ($DL_{\text{vanilla}}$), features selected by Multivariate Granger Causality ($DL_{\text{GC}}$), and features identified via PCMCI+ ($DL_{\text{PCMCI+}}$). The corresponding \(R^2\) values across forecast horizons are visualized in Figure~\ref{fig:R2_daily}.

For short lead times (1 month), $DL_{\text{vanilla}}$ achieves the lowest RMSE and highest \(R^2\), indicating superior performance when using all available features. However, as the forecast horizon extends, $DL_{\text{GC}}$ consistently outperforms other models at intermediate lead times (2, 4, and 5 months), showcasing the ability of MVGC-derived features to preserve long-range temporal dependencies. Conversely, $DL_{\text{PCMCI+}}$ excels at short-range forecasting, delivering the best MAE for 1-month and 3-month horizons, highlighting PCMCI+'s strength in isolating immediate causal drivers.

\begin{table}[h!]
\centering
\caption{Error metrics for daily models.}
\label{tab:daily_metrics_full}
\begin{tabular}{@{}ccccc@{}}
\toprule
\textbf{Lead Time} & \textbf{Metric} & $\mathbf{DL_{\textbf{vanilla}}}$ & $\mathbf{DL_{\textbf{GC}}}$ & $\mathbf{DL_{\textbf{PCMCI+}}}$ \\ \midrule
1-month  & RMSE (\%) & \textbf{7.777} & 8.017  & 8.043  \\
         & MAE (\%)  & 4.856  & 5.556  & \textbf{4.625} \\ \midrule
2-months & RMSE (\%) & 14.303 & \textbf{11.365} & 21.663 \\
         & MAE (\%)  & 7.003  & \textbf{5.593}  & 9.441  \\ \midrule
3-months & RMSE (\%) & \textbf{18.200} & 24.381 & 22.465 \\
         & MAE (\%)  & 7.719  & 8.883  & \textbf{8.524} \\ \midrule
4-months & RMSE (\%) & 13.708 & \textbf{11.274} & 13.779 \\
         & MAE (\%)  & 7.659  & \textbf{7.091}  & 8.163  \\ \midrule
5-months & RMSE (\%) & 20.340 & \textbf{17.586} & 20.822 \\
         & MAE (\%)  & 9.207  & \textbf{7.683}  & 10.294 \\ \midrule
6-months & RMSE (\%) & \textbf{15.342} & 15.038 & 17.573 \\
         & MAE (\%)  & 9.083  & \textbf{7.210}  & 8.979  \\ \bottomrule
\end{tabular}
\end{table}

\begin{figure}[h!]
\centering
\includegraphics[width=0.95\columnwidth]{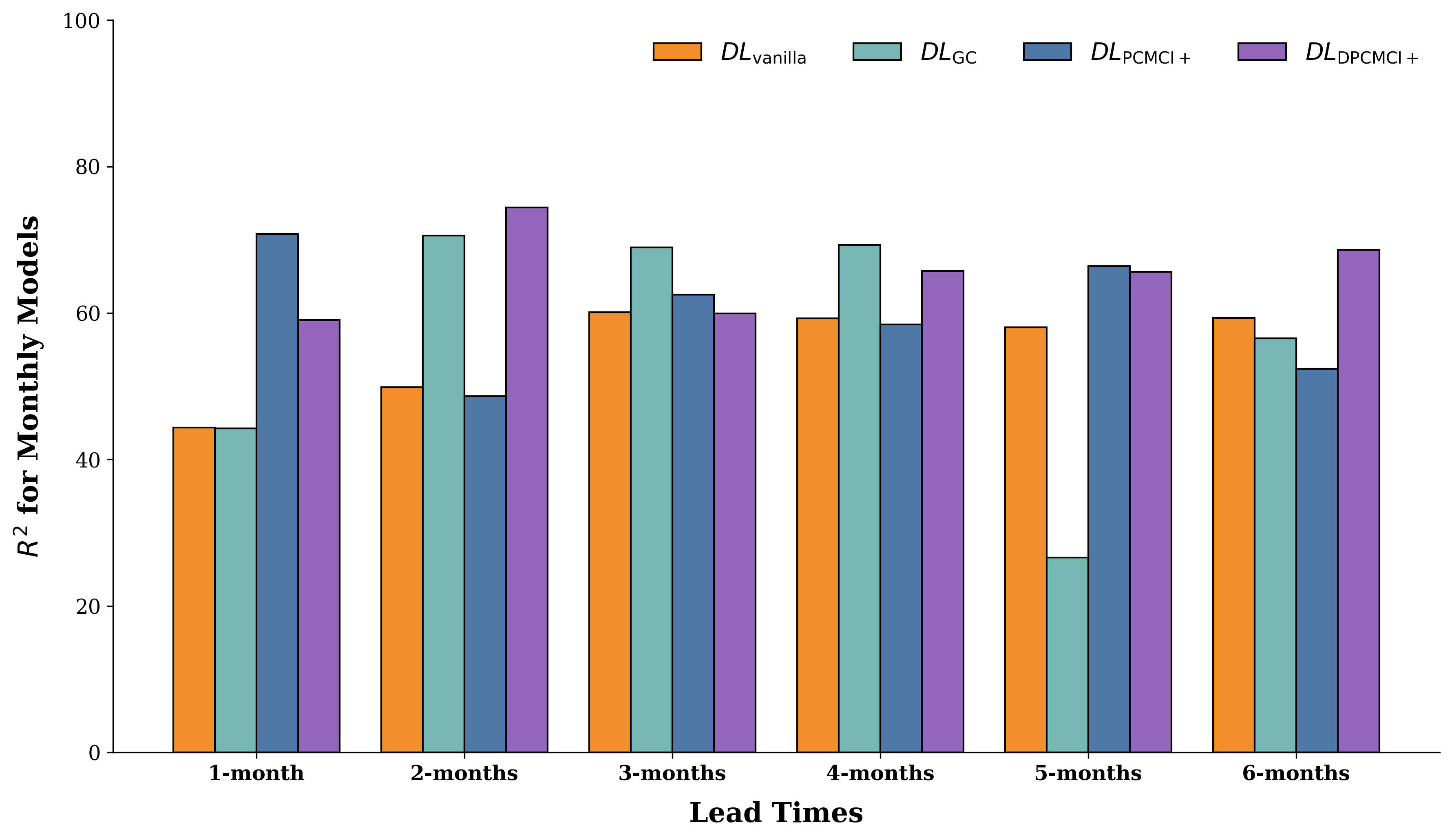}
\caption{\(R^2\) values for monthly models across the lead times.}
\label{fig:R2_monthly}
\end{figure}

For the monthly models, the results indicate generally higher prediction errors (Table~\ref{tab:monthly_metrics_full}) and lower \(R^2\) scores (Figure~\ref{fig:R2_monthly}) compared to their daily counterparts. This discrepancy is primarily due to the use of monthly-averaged input data, which reduces temporal granularity and limits the model’s capacity to capture short-term dynamics. 

Among the evaluated monthly configurations, $DL_{\text{PCMCI+}}$ delivers the best performance for 1-month forecasts, achieving the lowest RMSE and highest \(R^2\), affirming the value of PCMCI+-identified features for near-term prediction tasks. For intermediate lead times such as 2 months, $DL_{\text{DPCMCI+}}$ outperforms all other models. This model transfers causal features identified from daily data into the monthly forecasting context, effectively preserving high-resolution causal signals and improving generalization. This trend continues in extended lead times (5–6 months), where $DL_{\text{DPCMCI+}}$ consistently yields superior accuracy, underscoring the advantage of incorporating fine-scale causal features into longer-range forecasts. These findings emphasize the critical role of causally identified drivers in building predictive models that remain effective across different forecasting horizons.


\begin{table}[h!]
\centering
\caption{Error metrics for monthly models.}
\label{tab:monthly_metrics_full}
\resizebox{\columnwidth}{!}{%
\begin{tabular}{@{}cccccc@{}}
\toprule
\textbf{Lead Time} & \textbf{Metric} & $\mathbf{DL_{\textbf{vanilla}}}$ & $\mathbf{DL_{\textbf{GC}}}$ & $\mathbf{DL_{\textbf{PCMCI+}}}$ & $\mathbf{DL_{\textbf{DPCMCI+}}}$ \\ \midrule
1-month  & RMSE (\%) & 30.556 & 31.188 & \textbf{21.608} & 24.863 \\
         & MAE (\%)  & 16.169 & 15.839 & 16.884 & \textbf{15.839} \\ \midrule
2-months & RMSE (\%) & 27.081 & \textbf{23.451} & 26.040 & 19.851 \\
         & MAE (\%)  & 20.723 & 15.834 & 16.032 & \textbf{16.032} \\ \midrule
3-months & RMSE (\%) & 24.769 & \textbf{21.052} & 24.120 & 25.653 \\
         & MAE (\%)  & 20.156 & 18.397 & \textbf{18.783} & 19.676 \\ \midrule
4-months & RMSE (\%) & 23.007 & 22.675 & 25.133 & \textbf{22.105} \\
         & MAE (\%)  & 17.170 & \textbf{16.606} & 19.736 & 18.432 \\ \midrule
5-months & RMSE (\%) & 27.462 & 32.750 & \textbf{24.000} & 21.805 \\
         & MAE (\%)  & 19.522 & 18.324 & \textbf{17.835} & 18.949 \\ \midrule
6-months & RMSE (\%) & 27.815 & 27.247 & 26.971 & \textbf{21.883} \\
         & MAE (\%)  & 16.328 & 23.522 & 20.094 & \textbf{16.648} \\ \bottomrule
\end{tabular}%
}
\end{table}

Given that no single model consistently outperforms the others across all forecast horizons, the choice of model should be guided by the specific prediction objective. For short-term forecasting (e.g., 1-month lead time), $DL_{\text{PCMCI+}}$ demonstrates superior accuracy and is therefore recommended. At medium-range intervals, $DL_{\text{GC}}$ offers stronger generalization capabilities. For longer lead times (5–6 months), $DL_{\text{DPCMCI+}}$ capitalizes on high-resolution causal features derived from daily data, resulting in more stable and accurate predictions. These findings suggest that a hybrid modeling strategy—one that adapts model selection based on lead time—could further improve overall forecasting performance.

Despite the promising results, several challenges remain. Both PCMCI+ and MVGC are computationally demanding, particularly when applied to large causal graphs involving numerous environmental variables. Additionally, these methods exhibit sensitivity to hyperparameter choices, such as the selection of maximum lag, which can influence their robustness and scalability. Future work may address these limitations through algorithmic enhancements that reduce computational overhead~\cite{hossain2024time} and adopt adaptive strategies for lag selection, thereby improving the feasibility of deploying causality-informed models in operational Arctic sea ice forecasting pipelines.

\section{Conclusion}

This study presents a causality-informed deep learning framework for forecasting Arctic Sea Ice Extent (SIE), aiming to bridge the methodological gap between correlation-based machine learning techniques and causality-aware approaches. By applying Multivariate Granger Causality (MVGC) and PCMCI+, we systematically identify key ocean-atmospheric drivers with direct causal influence on SIE dynamics. This transition from statistical correlation to causal reasoning enables the model to focus on the most relevant variables, thereby reducing feature redundancy, streamlining architecture complexity, and improving computational efficiency by omitting non-informative inputs.

The proposed hybrid GRU-LSTM model demonstrates marked improvements in forecast accuracy when trained on causally selected features, compared to models that rely on the full correlated feature set. Empirical results across lead times from 1 to 6 months show consistent gains in predictive performance, evidenced by higher \(R^2\) values, and reductions in both RMSE and MAE.

Overall, this work underscores the advantages of incorporating causal discovery into deep learning workflows—leading to more interpretable, resilient, and computationally efficient models for dynamic, high-dimensional systems such as the Arctic climate. While this study focuses on sea ice prediction, the underlying framework is generalizable and holds promise for other application domains where causal understanding is critical for modeling complex temporal behaviors. Future directions will explore improving the scalability of causal discovery methods, implementing adaptive lag selection strategies, and extending the framework to account for spatial heterogeneity and multi-scale interactions.

\section*{Acknowledgments}
This work is supported by iHARP: NSF HDR Institute for Harnessing Data and Model Revolution in the Polar Regions (Award\# 2118285). The views expressed in this work do not necessarily reflect the policies of the NSF, and endorsement by the Federal Government should not be inferred.

\bibliographystyle{named}
\bibliography{References}

\begin{thebibliography}{}

\bibitem[\protect\citeauthoryear{Ali \bgroup \em et al.\egroup }{2021}]{ali2021sea}
Sahara Ali, Yiyi Huang, Xin Huang, and Jianwu Wang.
\newblock Sea ice forecasting using attention-based ensemble lstm.
\newblock {\em arXiv preprint arXiv:2108.00853}, 2021.

\bibitem[\protect\citeauthoryear{Andersson \bgroup \em et al.\egroup }{2021}]{andersson2021seasonal}
Tom~R Andersson, J~Scott Hosking, Mar{\'\i}a P{\'e}rez-Ortiz, Brooks Paige, Andrew Elliott, Chris Russell, Stephen Law, Daniel~C Jones, Jeremy Wilkinson, Tony Phillips, et~al.
\newblock Seasonal arctic sea ice forecasting with probabilistic deep learning.
\newblock {\em Nature communications}, 12(1):5124, 2021.

\bibitem[\protect\citeauthoryear{Barnett and Seth}{2014}]{barnett2014mvgc}
Lionel Barnett and Anil~K Seth.
\newblock The mvgc multivariate granger causality toolbox: a new approach to granger-causal inference.
\newblock {\em Journal of neuroscience methods}, 223:50--68, 2014.

\bibitem[\protect\citeauthoryear{Cavalieri \bgroup \em et al.\egroup }{1996}]{cavalieri1996sea}
Donald Cavalieri, Claire Parkinson, Per Gloersen, and H.~Zwally.
\newblock Sea ice concentrations from nimbus-7 smmr and dmsp ssm/i-ssmis passive microwave data, version 1, 1996.

\bibitem[\protect\citeauthoryear{Cho \bgroup \em et al.\egroup }{2014}]{cho2014learning}
Kyunghyun Cho, Bart Van~Merri{\"e}nboer, Caglar Gulcehre, Dzmitry Bahdanau, Fethi Bougares, Holger Schwenk, and Yoshua Bengio.
\newblock Learning phrase representations using rnn encoder-decoder for statistical machine translation.
\newblock {\em arXiv preprint arXiv:1406.1078}, 2014.

\bibitem[\protect\citeauthoryear{Dey \bgroup \em et al.\egroup }{2021}]{dey2021comparative}
Polash Dey, Emam Hossain, Md~Ishtiaque Hossain, Mohammed~Armanuzzaman Chowdhury, Md~Shariful Alam, Mohammad~Shahadat Hossain, and Karl Andersson.
\newblock Comparative analysis of recurrent neural networks in stock price prediction for different frequency domains.
\newblock {\em Algorithms}, 14(8):251, 2021.

\bibitem[\protect\citeauthoryear{Driscoll \bgroup \em et al.\egroup }{2024}]{driscoll2024data}
Simon Driscoll, Alberto Carrassi, Julien Brajard, Laurent Bertino, Einar {\'O}lason, Marc Bocquet, and Amos Lawless.
\newblock Data-driven emulation of melt ponds on arctic sea ice.
\newblock {\em EGUsphere}, 2024:1--18, 2024.

\bibitem[\protect\citeauthoryear{Dunmire \bgroup \em et al.\egroup }{2025}]{dunmireSGL2025}
Devon Dunmire, Aneesh~C. Subramanian, Emam Hossain, Md~Osman Gani, Alison~F. Banwell, Hammad Younas, and Brendan Myers.
\newblock Greenland ice sheet wide supraglacial lake evolution and dynamics: Insights from the 2018 and 2019 melt seasons.
\newblock {\em Earth and Space Science}, 12(2), 2025.

\bibitem[\protect\citeauthoryear{Ferdous \bgroup \em et al.\egroup }{2025}]{ferdous2025timegraph}
Muhammad~Hasan Ferdous, Emam Hossain, and Md~Osman Gani.
\newblock Timegraph: Synthetic benchmark datasets for robust time-series causal discovery.
\newblock In {\em Proceedings of the 31st ACM SIGKDD Conference on Knowledge Discovery and Data Mining V. 2}, pages 5425--5435, 2025.

\bibitem[\protect\citeauthoryear{Granger}{1969}]{granger1969investigating}
Clive~WJ Granger.
\newblock Investigating causal relations by econometric models and cross-spectral methods.
\newblock {\em Econometrica: journal of the Econometric Society}, pages 424--438, 1969.

\bibitem[\protect\citeauthoryear{Hasan \bgroup \em et al.\egroup }{2023}]{hasansurvey}
Uzma Hasan, Emam Hossain, and Md~Osman Gani.
\newblock A survey on causal discovery methods for iid and time series data.
\newblock {\em Transactions on Machine Learning Research}, 2023.

\bibitem[\protect\citeauthoryear{Hochreiter}{1997}]{hochreiter1997long}
S~Hochreiter.
\newblock Long short-term memory.
\newblock {\em Neural Computation MIT-Press}, 1997.

\bibitem[\protect\citeauthoryear{Hossain \bgroup \em et al.\egroup }{2020}]{hossain2020novel}
Emam Hossain, Mohd Arafath~Uddin Shariff, Mohammad~Shahadat Hossain, and Karl Andersson.
\newblock A novel deep learning approach to predict air quality index.
\newblock In {\em Proceedings of International Conference on Trends in Computational and Cognitive Engineering: Proceedings of TCCE 2020}, pages 367--381. Springer, 2020.

\bibitem[\protect\citeauthoryear{Hossain \bgroup \em et al.\egroup }{2024a}]{hossain2024incorporating}
Emam Hossain, Sahara Ali, Yiyi Huang, Nicole Schlegel, Jianwu Wang, Aneesh Subramanian, and Md~Osman Gani.
\newblock Incorporating causality with deep learning in predicting short-term and seasonal sea ice.
\newblock {\em AMS Annual Meeting}, 2024.

\bibitem[\protect\citeauthoryear{Hossain \bgroup \em et al.\egroup }{2024b}]{hossain2024time}
Emam Hossain, Md~Osman Gani, Devon Dunmire, Aneesh Subramanian, and Hammad Younas.
\newblock Time series classification of supraglacial lakes evolution over greenland ice sheet.
\newblock In {\em 2024 International Conference on Machine Learning and Applications (ICMLA)}. IEEE, 2024.

\bibitem[\protect\citeauthoryear{Hossain \bgroup \em et al.\egroup }{2025}]{hossain2025correlation}
Emam Hossain, Muhammad~Hasan Ferdous, Jianwu Wang, Aneesh Subramanian, and Md~Osman Gani.
\newblock Correlation to causation: A causal deep learning framework for arctic sea ice prediction.
\newblock In {\em 2025 IEEE International Conference on Pervasive Computing and Communications Workshops and other Affiliated Events (PerCom Workshops)}, pages 62--67. IEEE, 2025.

\bibitem[\protect\citeauthoryear{Islam and Hossain}{2021}]{islam2021foreign}
Md~Saiful Islam and Emam Hossain.
\newblock Foreign exchange currency rate prediction using a gru-lstm hybrid network.
\newblock {\em Soft Computing Letters}, 3:100009, 2021.

\bibitem[\protect\citeauthoryear{Kim \bgroup \em et al.\egroup }{2025}]{kim2025long}
Young~Jun Kim, Hyun-cheol Kim, Daehyeon Han, Julienne Stroeve, and Jungho Im.
\newblock Long-term prediction of arctic sea ice concentrations using deep learning: Effects of surface temperature, radiation, and wind conditions.
\newblock {\em Remote Sensing of Environment}, 318:114568, 2025.

\bibitem[\protect\citeauthoryear{Koo and Rahnemoonfar}{2024}]{koo2024hierarchical}
Younghyun Koo and Maryam Rahnemoonfar.
\newblock Hierarchical information-sharing convolutional neural network for the prediction of arctic sea ice concentration and velocity.
\newblock {\em IEEE Transactions on Geoscience and Remote Sensing}, 2024.

\bibitem[\protect\citeauthoryear{Li \bgroup \em et al.\egroup }{2024}]{li2024advancing}
Wenwen Li, Chia-Yu Hsu, and Marco Tedesco.
\newblock Advancing arctic sea ice remote sensing with ai and deep learning: now and future.
\newblock {\em EGUsphere}, 2024:1--36, 2024.

\bibitem[\protect\citeauthoryear{Liu \bgroup \em et al.\egroup }{2024}]{liu2024physics}
Quanhong Liu, Yangjun Wang, Ren Zhang, Lujun Zhang, Hengqian Yan, and Kefeng Liu.
\newblock Physics-informed deep convolutional network for combined sea ice concentration and velocity prediction.
\newblock {\em Ocean Engineering}, 313:119440, 2024.

\bibitem[\protect\citeauthoryear{Oliveira \bgroup \em et al.\egroup }{2024}]{oliveira2024causality}
Daniel~Cunha Oliveira, Yutong Lu, Xi~Lin, Mihai Cucuringu, and Andre Fujita.
\newblock Causality-inspired models for financial time series forecasting.
\newblock {\em arXiv preprint arXiv:2408.09960}, 2024.

\bibitem[\protect\citeauthoryear{Pearl and Mackenzie}{2018}]{pearl2018book}
Judea Pearl and Dana Mackenzie.
\newblock {\em The book of why: the new science of cause and effect}.
\newblock Basic books, 2018.

\bibitem[\protect\citeauthoryear{Ren \bgroup \em et al.\egroup }{2024}]{ren2024sicnet}
Yibin Ren, Xiaofeng Li, and Yunhe Wang.
\newblock Sicnet season v1. 0: a transformer-based deep learning model for seasonal arctic sea ice prediction by integrating sea ice thickness data.
\newblock {\em Geoscientific Model Development Discussions}, 2024:1--20, 2024.

\bibitem[\protect\citeauthoryear{Runge}{2020}]{runge2020discovering}
Jakob Runge.
\newblock Discovering contemporaneous and lagged causal relations in autocorrelated nonlinear time series datasets.
\newblock In {\em Conference on Uncertainty in Artificial Intelligence}, pages 1388--1397. Pmlr, 2020.

\bibitem[\protect\citeauthoryear{Xu \bgroup \em et al.\egroup }{2024}]{xu2024sifm}
Jingyi Xu, Yeqi Luo, Weidong Yang, Keyi Liu, Shengnan Wang, Ben Fei, and Lei Bai.
\newblock Sifm: A foundation model for multi-granularity arctic sea ice forecasting.
\newblock {\em arXiv preprint arXiv:2410.14732}, 2024.

\bibitem[\protect\citeauthoryear{Zhu \bgroup \em et al.\egroup }{2024}]{zhu2024stdnet}
Xu~Zhu, Jing Wang, Guojun Wang, Yangming Jiang, Yi~Sun, and Huihui Zhao.
\newblock Stdnet: Spatio-temporal decompose network for predicting arctic sea ice concentration.
\newblock {\em Remote Sensing}, 16(23):4534, 2024.

\end{thebibliography}

\end{document}